\documentclass{INTERSPEECH2023}

\interspeechcameraready

\usepackage{times}
\usepackage{latexsym}
\usepackage{multirow, tabu}
\usepackage{float}
\usepackage{graphicx}
\usepackage{placeins}
\usepackage{xcolor}
\usepackage{ulem}
\usepackage{dashrule}
\usepackage{hyperref}
\usepackage{arydshln}

\newcommand{\sect}[1]{Section~\ref{sec:#1}}

\newcommand{\fig}[1]{Figure~\ref{fig:#1}}

\newcommand{\tbl}[1]{Table~\ref{tab:#1}}

\newcommand{\ignore}[1]{}

\makeatletter
\DeclareRobustCommand\onedot{\futurelet\@let@token\@onedot}
\def\@onedot{\ifx\@let@token.\else.\null\fi\xspace}
\makeatother

\definecolor{MyDarkBlue}{rgb}{0,0.08,1}
\definecolor{MyDarkGreen}{rgb}{0.02,0.6,0.02}
\definecolor{MyDarkRed}{rgb}{0.8,0.02,0.02}
\definecolor{MyDarkOrange}{rgb}{0.40,0.2,0.02}
\definecolor{MyPurple}{RGB}{111,0,255}
\definecolor{MyRed}{rgb}{1.0,0.0,0.0}
\definecolor{MyGold}{rgb}{0.75,0.6,0.12}
\definecolor{MyDarkgray}{rgb}{0.66, 0.66, 0.66}

\colorlet{Delete}{MyDarkRed}
\colorlet{Replace}{MyDarkBlue}
\colorlet{Replaced}{MyDarkGreen}

\makeatletter
  \UL@protected\def\delline{\bgroup\markoverwith{\textcolor{Delete}{\rule[0.5ex]{2pt}{0.8pt}}}\ULon}

  \UL@protected\def\substituteuline{\bgroup\markoverwith{\textcolor{Replace}{\rule[-0.6ex]{2pt}{0.8pt}}}\ULon}

  \UL@protected\def\substituteoline{\bgroup\markoverwith{\textcolor{Replace}{\rule[1.6ex]{2pt}{0.8pt}}}\ULon}

  \UL@protected\def\replaceduline{\bgroup\markoverwith{\textcolor{Replaced}{\rule[-0.6ex]{2pt}{0.8pt}}}\ULon}

  \UL@protected\def\replacedoline{\bgroup\markoverwith{\textcolor{Replaced}{\rule[1.6ex]{2pt}{0.8pt}}}\ULon}
    
  \UL@protected\def\substituteulineprev{\leavevmode \bgroup 
    \UL@setULdepth
    \ifx\UL@on\UL@onin \advance\ULdepth2\p@\fi
    \markoverwith{\begingroup
       \lower\ULdepth\hbox{\kern.06em \textcolor{Replace}{.}\kern.04em}%
       \endgroup}%
    \ULon}

      \UL@protected\def\substitutedotuline{\leavevmode \bgroup 
    \UL@setULdepth
    \ifx\UL@on\UL@onin \advance\ULdepth2\p@\fi
    \markoverwith{\begingroup
       \lower\ULdepth\hbox{\kern.06em \textcolor{Replace}{\LARGE{.}}\kern.04em}%
       \endgroup}%
    \ULon}
    
  \UL@protected\def\replacedulineprev{\leavevmode \bgroup 
    \UL@setULdepth
    \ifx\UL@on\UL@onin \advance\ULdepth2\p@\fi
    \markoverwith{\kern.13em
    \vtop{\color{Replaced}\kern\ULdepth \hrule height .1em width .3em}%
    \kern.13em}\ULon}

      \UL@protected\def\replaceddashuline{\leavevmode \bgroup 
    \UL@setULdepth
    \ifx\UL@on\UL@onin \advance\ULdepth2\p@\fi
    \markoverwith{\kern.13em
    \vtop{\color{Replaced}\kern\ULdepth \hrule height .2em width .3em}%
    \kern.13em}\ULon}
    
  \UL@protected\def\substituteolineprev{\leavevmode \bgroup 
    \UL@setULdepth
    \ifx\UL@on\UL@onin \advance\ULdepth2\p@\fi
    \markoverwith{\begingroup
       \lower\ULdepth\hbox{\kern.06em \textcolor{Replace}{.}\kern.04em}%
       \endgroup}%
    \ULon}
    
  \UL@protected\def\replacedolineprev{\leavevmode \bgroup 
    \UL@setULdepth
    \ifx\UL@on\UL@onin \advance\ULdepth2\p@\fi
    \markoverwith{\kern.13em
    \vtop{\color{Replaced}\kern\ULdepth \hrule height .1em width .3em}%
    \kern.13em}\ULon}
\makeatother

\newcommand{\deletecolor}[1]{\textcolor{Delete}{#1}} 
\newcommand{\substitutecolor}[1]{\textcolor{Replace}{#1}} 
\newcommand{\replacedcolor}[1]{\textcolor{Replaced}{#1}} 

\title{5IDER: Unified Query Rewriting for Steering, Intent Carryover, Disfluencies, Entity Carryover and Repair}

\name{Jiarui Lu*\thanks{*Equal contribution}, Bo-Hsiang Tseng*, Joel Ruben Antony Moniz*,\\ Site Li, Xueyun Zhu, Hong Yu, Murat Akbacak}
\address{
  Apple}
\email{\{jiarui\_lu,bohsiang\_tseng,joelrubenantony\_moniz,\\site\_li,xueyun\_zhu,hong\_yu,makbacak\}@apple.com}

\begin{document}

\maketitle

\begin{abstract}
Providing voice assistants the ability to navigate multi-turn conversations is a challenging problem. Handling multi-turn interactions requires the system to understand various conversational use-cases, such as steering, intent carryover, disfluencies, entity carryover, and repair. The complexity of this problem is compounded by the fact that these use-cases mix with each other, often appearing simultaneously in natural language. This work proposes a non-autoregressive query rewriting architecture that can handle not only the five aforementioned tasks, but also complex compositions of these use-cases. We show that our proposed model has competitive single task performance compared to the baseline approach, and even outperforms a fine-tuned T5 model in use-case compositions, despite being 15 times smaller in parameters and 25 times faster in latency.
\end{abstract}


\section{Introduction}

With voice assistants making large improvements recently, natural conversations with virtual assistants has become a more common expectation. Users tend to interact with virtual assistant in an increasingly contextual fashion, expecting them to behave more like a human agent \cite{kiseleva2016understanding}. This brings out a growing challenge: the understanding system behind a voice assistant needs to be robust to several aspects of natural, conversational language that have been traditionally difficult to deal with.

One line of challenges involves the ability to use context from one turn to complete the understanding of another in multi-turn conversations. This includes use-cases like \textbf{steering} (where a follow-up turn is used to provide clarifying information to a previous turn), \textbf{intent carryover} (where a follow-up turn implicitly has the same intent as a previous turn, but for a different entity) \cite{su-etal-2019-improving} and \textbf{entity carryover} (where a follow-up turn refers to an entity in a previous turn, often through anaphora or nominal ellipses) \cite{maqbool2022zero, yang-etal-2019-end-end-neural, vakulenko2021question, rastogi2019scaling}. Another set of challenges involves artifacts in human speech that appear by virtue of humans changing their mind during a conversation, or beginning a conversation without having fully decided their intent or how they would like to convey it. This yields \textbf{disfluencies} (i.e., artifacts in speech such as filler words that don't contribute to the intent of a query and can be removed) and \textbf{repair} (where a user corrects the intent and/or entity previously referenced) \cite{chen2022teaching}. This problem of understanding spoken language is further complicated by the ability of these artifacts to mix with each other, with these phenomena often appearing simultaneously in natural language. We illustrate these problems with examples in Table~\ref{tab:usecase_examples}. The 5 aforementioned challenges are shown at the top, with a pair of context and follow-up queries. We also present 2 compositional challenges at the bottom, where multiple conversational challenges are involved.

    \begin{table*}[!ht]
        \caption{Examples of conversational use cases and their compositions tackled in this work. Examples shown are author-created queries based on anonymized and randomly sampled virtual assistant logs. Tokens marked in \deletecolor{red} indicate those marked for deletion. Tokens marked in \replacedcolor{green} or with a \replaceduline{green underline} are those marked to be replaced, while those marked in \substitutecolor{blue} or with a \substituteuline{blue underline} are the detected replacements. Note that colorization indicates a first step substitution, and underlines indicate a second substitution step (which only ever occurs in compositional use cases).}
  \label{tab:usecase_examples}
    \vspace{-0.5em}
  \centering
  \resizebox{\linewidth}{!}{%
  \begin{tabu}{l|c|c|c|c}
  \tabucline [1pt]{1}
  \textbf{Use Case}    & \textbf{Context} &  & \textbf{Follow-up} & \textbf{Rewrite} \\ \hline \hline
  \textbf{Steering} & Play Sweeny Todd & \deletecolor{[SEP]} & In my living room & Play Sweeny Todd in my living room \\ 
  \textbf{Intent Carryover} & How old is \replacedcolor{Homer Simpson} & \deletecolor{[SEP]} & \deletecolor{What about} \substitutecolor{Bart Simpson} & How old is Bart Simpson \\ 
  \textbf{Disfluency} & - & [SEP] & Take me to \replacedcolor{Suki Sushi} \deletecolor{no I said} \substitutecolor{Fuki Sushi} & Take me to Fuki Sushi \\ 
  \textbf{Entity Carryover} & When does \substitutecolor{Rocket Sushi} close & [SEP] & How long does it take to drive \replacedcolor{there} & How long does it take to drive to Rocket Sushi \\ 
  \textbf{Repair} & How far is \replacedcolor{San Jose} by car & \deletecolor{[SEP]} &  \deletecolor{I meant} \substitutecolor{San Francisco} & How far is San Francisco by car \\ \hline
  \textbf{Entity Carryover + Intent Carryover} & How tall is \replaceduline{\substitutecolor{Homer Simpson}} & \deletecolor{[SEP]} & \deletecolor{What about} \substituteuline{\replacedcolor{his} wife} & How tall is Homer Simpson's wife \\ 
  \textbf{Entity Carryover + Repair} & Who is \replaceduline{\substitutecolor{Homer Simpson's} eldest doctor} & \deletecolor{[SEP]} & \deletecolor{I said} \substituteuline{\replacedcolor{his} eldest daughter} & Who is Homer Simpson's eldest daughter \\ 
  \tabucline [1pt]{1}
  \end{tabu}%
  }
  \vspace{-1.25em}
  \end{table*}

Traditional approaches handle each of these challenges with dedicated systems: entity carryover is often tackled with coreference \cite{ng2002improving, lee2017end} and ellipsis \cite{carbonell-1983-discourse} resolution; intent carryover can be treated as a form of slot carryover \cite{Naik2018, chen-etal-2019-improving-long}; disfluencies can be solved as a sequence tagging problem \cite{zayats16_interspeech}. While each dedicated system offers competitive performance for each challenge, a solution for all challenges would involve cascading these systems, which introduces error propagation, extra latency and disk space, and management complexities. A unified problem formulation could yield a single, joint modeling solution. 

One such formulation is query rewriting  \cite{nguyen2021user, quan-etal-2019-gecor, 10.1145/3397271.3401323, Tseng2021CREADCR}, which reformulates contextual queries into their context-independent counterparts. Examples of target rewritten queries for all use cases and their combinations are shown in the \textit{Rewrite} column in Table~\ref{tab:usecase_examples}. 
For example, for the composition of Entity Carryover and Repair, given the context query \textit{``Who is Homer Simpson's eldest doctor''} and the follow-up query \textit{``I said his eldest daughter''}, the reference to the context entity \textit{Homer Simpson} needs to be resolved along with a repair of the ASR error; solving both then yields a self-contained query: \textit{``Who is Homer Simpson's eldest daughter''}. 
While entity carryover \cite{quan-etal-2019-gecor, Tseng2021CREADCR}, intent carryover \cite{su-etal-2019-improving, 10.1145/3397271.3401323} (and their combinations), disfluency \cite{gupta-etal-2021-disfl} and repair \cite{nguyen2021user} have been studied as query rewriting problems individually, to the best of our knowledge, this is the first work to look at all 5 challenges (including the under-studied challenge of handling steering) jointly, and to investigate compositional challenges.

Many prior query rewriting works consider it a summarization task \cite{su-etal-2019-improving, quan-etal-2019-gecor,   10.1145/3397271.3401323, Tseng2021CREADCR, mallinson2022edit5, radford2019language}. While these models usually offer competitive performance, they suffer from heavy latency costs brought by the decoding loop, which would have a significant impact on the responsiveness of a virtual assistant. 
To mitigate this, text-editing rewrite models were proposed, which output a sequence of edit actions (like deletion, insertion, swap and reordering) instead of generating tokens. 
Applying these edits on the source tokens yields the rewritten query, drastically reducing the output search space. 
In LaserTagger \cite{malmi-etal-2019-encode}, the authors use a single layer Transformer Decoder paired with a BERT encoder \cite{devlin-etal-2019-bert} to reduce decoding latency. FELIX \cite{mallinson-etal-2020-felix} goes one step further using a non-autoregressive model with tagging, reordering and inserting, eliminating the decoding loop completely.
 
 In this paper, we aim to address this latency issue with a non-autoregressive edit-based model using composable edit-actions. In addition, we evaluate the query rewriting problem in a multitask setting, including compositional use cases, which, to the best of our knowledge, is a first. Through experimental evaluation, we show that our proposed approach has competitive single- and multi-task performance compared to baseline approaches on the target use cases, and can even surpass a fine-tuned T5-small \cite{raffel2020exploring} model in use-case combinations, despite being 15x smaller in parameters. Our main contributions are:
\begin{enumerate}
    \item We propose a joint edit-based query rewrite model to handle all five conversational understanding use cases and their compositions.
    \item We show that our non-autoregressive model achieves performance close to or even better than strong baselines, with 25x faster latency, 15x smaller disk size and 20x better compositional data efficiency.
\end{enumerate}

\section{Data}

Collecting data for conversational use cases that are not currently supported  by the virtual assistant is challenging because of the cold-start problem: users' behavior attunes to the known capability of the virtual assistant, and interactions targeting unsupported use cases rarely exist. As a result, collecting such data requires clever strategies and annotations, or partial synthesis. Our data collection process starts by randomly sampling virtual assistant logs from anonymized opt-in users, which we use to create 6 datasets in total, 5 for each of the tasks we focus on, and 1 containing compositions of the tasks, all of which are either human annotated or synthetically created.

Our entity and intent carryover datasets start with identifying consecutive query pairs with common entities or intents between them. Given these pairs, annotators provide a contextual query that simplifies the follow-up through entity or intent carryover, similar to  \cite{10.1145/3397271.3401323}. 

For our repair dataset, when users correct themselves by manually editing ASR transcriptions, we extract the slot where the correction was performed. We then use the slot to populate synthetic templates such as (original turn, correction phrase ("No I meant", etc.) + slot, correction turn), where the first element represents the context turn, the second element represents the followup-turn, and the third element represents the rewrite.

We formulate our disfluency dataset similarly to our repair dataset above, but use synthetic templates to stitch together the correction slot inside the query, randomly adding interregnum to simulate the disfluency. 

For steering, we look at consecutive queries in which the first query is an exact prefix of the second. The non-prefix part of the second query helps us simulate what a user who wishes to use a follow-up in a steering fashion might say.

Our compositional dataset is created through synthetic templates because their rarity makes collecting real-world data exceedingly difficult. Each template requires two of the aforementioned tasks to be resolved. We identified 5 valid challenge pairs, from which we created 20 unique templates. Templates used to create training, validation and test sets are disjoint to keep this task challenging. Random entities are then filled into the templates to create the data.

Each of the 5 datasets contains 60k datapoints, and the compositional dataset contains 20k datapoints. Each dataset has an 8:1:1 training:validation:test split. On average, context turns have 5.6 tokens, current turns have 5.2 tokens, rewritten turn have 6.0 tokens, and 47\% of tokens in the rewritten turn can only be found in the context turn.

\section{5IDER}

\begin{figure*}[t]
    \centering
    \includegraphics[width=1.0\linewidth]{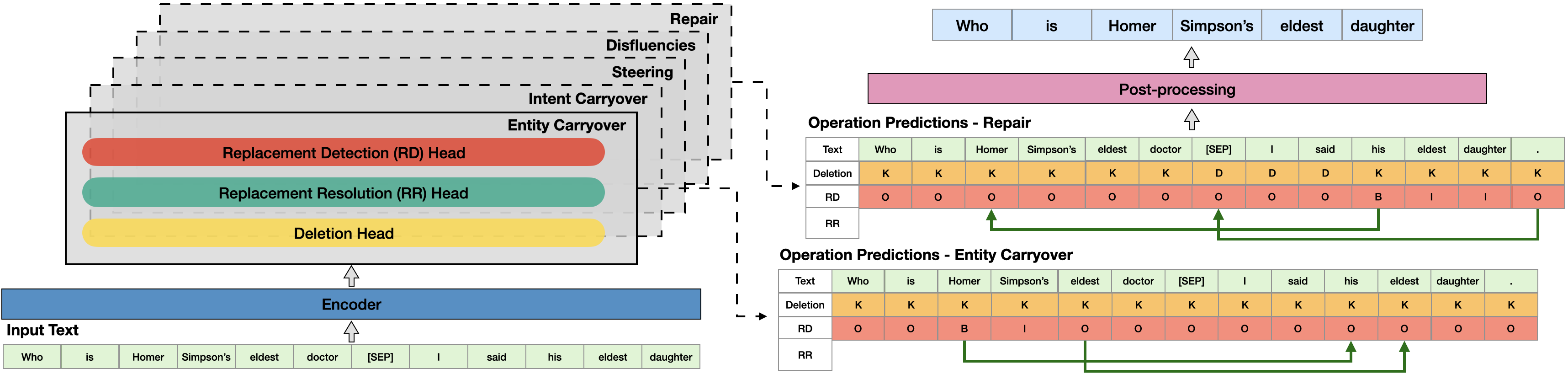}
    \caption{The proposed 5IDER architecture. The input text, the concatenation of two turns, is first encoded with TinyBert and an LSTM encoder. Then, 5 edit operation predictors (one for each task), perform Replacement Detection, Replacement Resolution and Deletion Detection for each task. Finally, all edit operations are applied in a post-processing step.}
    \label{fig:proposed_model}
    \vspace{-1.0em}
\end{figure*}

Our proposed method 5IDER (shown in \fig{proposed_model}) is an encoder-based model that learns to predict the defined \textit{edit operations} to accomplish rewriting across all use cases. In this section, we first define these edit operations, then explain how our model architecture and training objective help to learn these operations.

\subsection{Rewriting with edit operations}
\label{sec:edit_ops}
To accomplish rewriting across all different use cases, it is essential to capture the relationship between text spans within the input sequence. For instance, in the example of \textit{entity carryover} shown in \tbl{usecase_examples}, the model needs to resolve the pronoun \textit{it} in the follow-up turn to the named entity \textit{Rocket Sushi} mentioned in the context. To achieve this goal, we define two edit operations, substitution and deletion, which are made possible by three model components:

\begin{itemize}
    \item Replacement detection: this component detects the replacement, a text span which will substitute another text span to make a self-contained rewrite.
    \item Replacement resolution: this component detects the replaced, a text span to be substituted by the replacement.
    \item Deletion: this component detects tokens that need to be deleted to complete a rewrite.
\end{itemize}

These edit operations act on the concatenated dialog history, in the form of \textit{context query} + \textit{separator token ([SEP])} + \textit{follow-up query}.
For single use case data, we apply the following post-processing steps to create the rewritten query: first, apply deletion; second, apply substitution by deleting the replacement, and substituting the replaced with the replacement; finally, extract the token sequence starting from the last [SEP] to the end of the sequence (or the whole token sequence if no [SEP]s are left). As an example, the replacement, replaced and deletion for each of the 5 use cases are color coded in blue, green and red in Table~\ref{tab:usecase_examples}. This thus implies that, as in \cite{su-etal-2019-improving}, the output vocabulary of this design is limited to the input tokens.

This edit operation design differs from \cite{malmi-etal-2019-encode, mallinson-etal-2020-felix}: instead of general purpose edit operations, each use case has an independent interpretation of the operations. For entity carryover, a substitution is akin to anaphora resolution, while for repair, a substitution somewhat resembles entity linking. 
The motivation behind this is to create a model architecture with the ability to apply edit operations independently for each use case, automatically solving use case composition by simply composing these operations.
An example is shown in Figure~\ref{fig:proposed_model}. In \textit{Who is Homer Simpson's eldest doctor [SEP] I said his eldest daughter}, entity carryover requires \textit{his} being replaced by \textit{Homer Simpson's}, repair requires \textit{Homer Simpson's eldest doctor} being replaced by \textit{his eldest daughter}, and \textit{[SEP] I said} being deleted. To tackle this, we need to modify the second step of our aforementioned post-processing logic. We define the concept of \textbf{substitution dependency}. A substitution is dependent on another, if its replacement contains the replaced of the other. In our example, the repair substitution is dependent on the entity carryover substitution. 
We perform a topological sort for substitutions, applying those without dependencies and moving upwards. We thus first apply the entity carryover substitution to obtain \textit{Who is eldest doctor [SEP] I said Homer Simpson's eldest daughter}. 
After this, the repair's replaced has to be adjusted from \textit{Homer Simpson's eldest doctor} to \textit{eldest doctor}. 
We then apply the repair's substitution to obtain \textit{Who is Homer Simpson's eldest daughter [SEP] I said}. Finally, after deletion, we have the rewritten query \textit{Who is Homer Simpson's eldest daughter}. 

\subsection{Model Architecture}
As shown in \fig{proposed_model}, 5IDER takes as input the concatenation of text from the context and the follow-up turn.
The input sequence is then passed through a frozen TinyBert \cite{jiao-etal-2020-tinybert} model to obtain contextualized embeddings of input tokens, which is then further encoded with a trainable BiLSTM encoder.

To handle data with an arbitrary mixture of use cases, the model contains 5 copies of the 3 components mentioned in \sect{edit_ops}, one for each use case.
First, the \textbf{Replacement Detection} (RD) head predicts which span in the input is a replacement. This is modeled as a BIO sequence tagging task. As shown in the running example, the entity \textit{``Homer Simpson's''} is identified as the replacement in \textit{entity carryover}, and \textit{``his eldest daughter''} in the \textit{repair} use case. When no valid replacements are predicted in a use case component (i.e., the output of the RD head is $O$ across the input sequence), the input sequence does not require any substitutions for that use case (e.g., use cases \textit{intent carryover}, \textit{steering} and \textit{disfluencies} in the running example).
Second, the \textbf{Replacement Resolution} (RR) head identifies the text that the replacement needs to substitute (i.e., the replaced text). Concretely, biaffine attention \cite{dozat2016deep} is employed to perform self-attention to capture the relationship between input tokens.
Similar to the mechanism of locating the answer span in machine comprehension, the attention distribution at the position of the beginning (end) of the replacement is supervised to attend to the beginning (end) of the corresponding replaced, as shown by the solid green arrows.
Finally, the \textbf{Deletion} head, a binary classifier, is applied at each position to determine whether to delete ($D$) or keep ($K$) the token. The model predicts these edit operations separately for each use case, and they are then consolidated to form the final rewrite, as described in \sect{edit_ops} above.

\subsection{Optimization}
When optimizing the model, supervision is provided for all three edit operation heads.
For \textbf{Replacement Detection}, RD heads across all use cases are optimized with the loss $L^{RD}$, using cross-entropy (CE) between the ground-truth and the predicted RD sequence:
\begin{equation}
    L^{RD} = \frac{1}{|U|}\sum^{|U|}_{u}\sum_{i}^{T} CE(p^{RD}_{u, i}, y^{RD}_{u, i})
\end{equation}
where $u$ and $i$ are respectively the indices for the use case and input position. $U$ is the total number of use cases; $T$ is the length of the input sequence; $p^{RD}_{u, i} \in \mathbb{R}^{3}$ is the BIO prediction at the position $i$ for the use case $u$.

For \textbf{Replacement Resolution}, the RR heads are supervised only within use cases where a replacement exists in the input sequence, in which case the loss is defined as the sum of the cross-entropy between the ground-truth and the predictions at the positions of the replacement boundary (i.e., start \& end):
\begin{equation}
    L^{RR}_{u} = 
\begin{cases}
    \sum_{i=\{start, end\}} CE(p^{RR}_{u, i}, y^{RR}_{u, i}),&\text{ \hspace{-2.5mm}replacement exists}\\
    0, & \text{\hspace{-2.0mm}otherwise}
\end{cases}
\end{equation}
where $p^{RR}_{u, i} \in \mathbb{R}^{T}$ is the prediction over the input sequence. The overall RR loss is added up across use cases: \\
$L^{RR} = \sum_{u}^{|U|} L^{RR}_{u}$.

For \textbf{Deletion}, the deletion heads across all use cases are trained with a binary signal for each input token:
\begin{equation}
    L^{Del} = \frac{1}{|U|} \sum^{|U|}_{u} \sum_{i}^{T} CE(p^{Del}_{u, i}, y^{Del}_{u, i})
\end{equation}
where $p^{Del}_{u, i} \in \mathbb{R}^{2}$ is the binary prediction on whether to keep or delete the token at position $i$ for the use case $u$.

The overall loss, $L$, used for model training is the sum of all three losses with equal weight: $L = L^{RD} + L^{RR} + L^{Del}$.

\section{Experimental Setup}

\textbf{Hyperparameters} For 5IDER, we use a 192-dim frozen TinyBERT embedding \cite{jiao-etal-2020-tinybert} and 128-dim for all LSTM hidden states, biaffine attention RR heads, classifiers in RD, and deletion heads. During training, the model is trained with batch size 64, learning rate 6e-4 with Adam and dropout of 0.2.

\textbf{Baseline Systems}
We compare 5IDER with baseline models to show the efficacy of the designed editing mechanism in our model. The first baseline is a BiLSTM-based seq2seq model\footnote{We also experimented with similar sized Transformer variants, and found that even with 3 times more parameters, Transformers still perform worse than an LSTM.} with a copy mechanism \cite{gu2016incorporating}. This model is close to our system in terms of model capacity: with the same set of hyperparameters, its model size is approximately twice that of 5IDER due to the additional decoder. We also compare against generative (T5-Small 
 \cite{raffel2020exploring}) and edit-based (LaserTagger \cite{malmi-etal-2019-encode} and FELIX \cite{mallinson-etal-2020-felix}) transformer models to see if the rewriting performance can be boosted at the cost of run-time speed.
The input to the two generation baselines are the same as 5IDER.
All baselines are fine-tuned and optimized across all use cases.

\begin{table}[t]
\caption{Model comparison in model size, disk size and latency. Latency is measured on the same hardware and averaged across test sets. Seq2Seq latency is set as 1 unit for ease of comparison. }
\label{tab:model_comparison}
\vspace{-0.5em}
\centering
\resizebox{0.8\linewidth}{!}{%
\begin{tabu}{lccc}
\tabucline [1pt]{1}
Model    & No. Parameters & Disk Size & Relative Latency  \\ \hline \hline
Seq2Seq & 4.5 M      & 17.9 Mb       & 1 x    \\ 
LaserTagger & 110 M & 430 Mb & 25.1 x \\ 
Felix       &  220 M & 933 Mb & 22 x \\
T5-Small       & 60.5 M       & 242 Mb     &  11.3 x       \\ 
5IDER & 4.2M & 16.9 Mb & 0.4 x \\ 

\tabucline [1pt]{1}
\end{tabu}%
}
\vspace{-0.3em}
\end{table}

\begin{table}[t!]
\caption{Exact match accuracy (\%) on different use cases and their average under two training setups.}
\label{tab:result_single}
\centering
\vspace{-0.5em}
\resizebox{1.0\linewidth}{!}{%
\begin{tabu}{lccccc|c}
\tabucline [1pt]{1}
Model    & Intent & Entity & Repair & Disfluency & Steering & Avg. \\ \hline
\multicolumn{6}{c}{Singe-task training} \\ \hline
Seq2Seq & 94.9 & 84.4 &  78.0 & 73.1 & 98.2 & 85.7 \\
LaserTagger & 84.7 & 49.3 & 64.6 & 63.0 & 99.9 & 72.3 \\
Felix & 96.0 & 83.7 & 85.6 & 77.4 & 97.1 & 88.0  \\
T5-Small    & 96.3 & 90.6 & 83.5 & 82.0 & 95.1 & 89.5  \\ 
5IDER (ours) & 95.5 & 86.5 & 80.2 & 79.2 & 100.0 & 88.3\\
\hline
\multicolumn{6}{c}{Multi-task training} \\ \hline
Seq2Seq & 95.4 & 84.5 & 78.9 & 75.6 & 97.1 & 85.3  \\
LaserTagger & 84.5 & 49.1 & 62.9 & 60.0 & 99.8 & 71.3 \\
Felix  & 88.1 & 1.7 & 80.9 & 25.0 & 44.1 & 48.0 \\
T5-Small       & 95.7 & 89.3 & 83.3 & 83.6 & 93.1 & 89.0  \\ 
5IDER (ours) & 95.3 & 86.1 & 81.0 & 80.6 & 98.9  & 88.4 \\
\tabucline [1pt]{1}
\end{tabu}%
}
\vspace{-1.2em}
\end{table}

\section{Results}

\subsection{Inference Time} 
We compare the models in terms of model size, disk size and latency in \tbl{model_comparison}. As shown, 5IDER has the best latency and disk size out of the five, occupying less than 20Mb of disk space and being over 25x faster than T5-Small, LaserTagger and Felix. 

\subsection{Single-Task and Multi-Task Experiments}

We test the models' capacity by training and testing on the same use case (single-task training), as shown in the first part of Table~\ref{tab:result_single}. In general, all models perform similarly well across tasks, except for LaserTagger, which struggles on some use cases.\footnote{LaserTagger's edit operations include swapping sentence order, deletion and insertion with a limited vocabulary. It thus cannot consistently support use cases like entity carryover, which require a text span from the context query to be copied to the middle of the follow-up.} The general trend of good performance indicates that single-task training is straightforward for these models.

The more challenging but practical setup is the multi-task setting, where the same model needs to handle various linguistic patterns. Results are shown in the second part of Table~\ref{tab:result_single}. Although the task difficulty increases, 
5IDER is still capable of tacking all 5 use cases, performing competitively with the strong T5 baseline (with only a 0.6\% gap on average) in spite of its 25x latency and disk size benefit.
However, the other two edit-based models perform much worse than 5IDER, despite using a BERT base encoder. Among them, Felix is unable to learn the different rewriting patterns in joint training (with an accuracy of 1.7\% on the Entity use case).
This indicates that our edit-based model is not only light-weight and low-latency, but also good at generalization in a multitask setting.

\subsection{Few-shot Use Case Composition}

\begin{figure}
  \centering
  \includegraphics[width=0.85\linewidth]{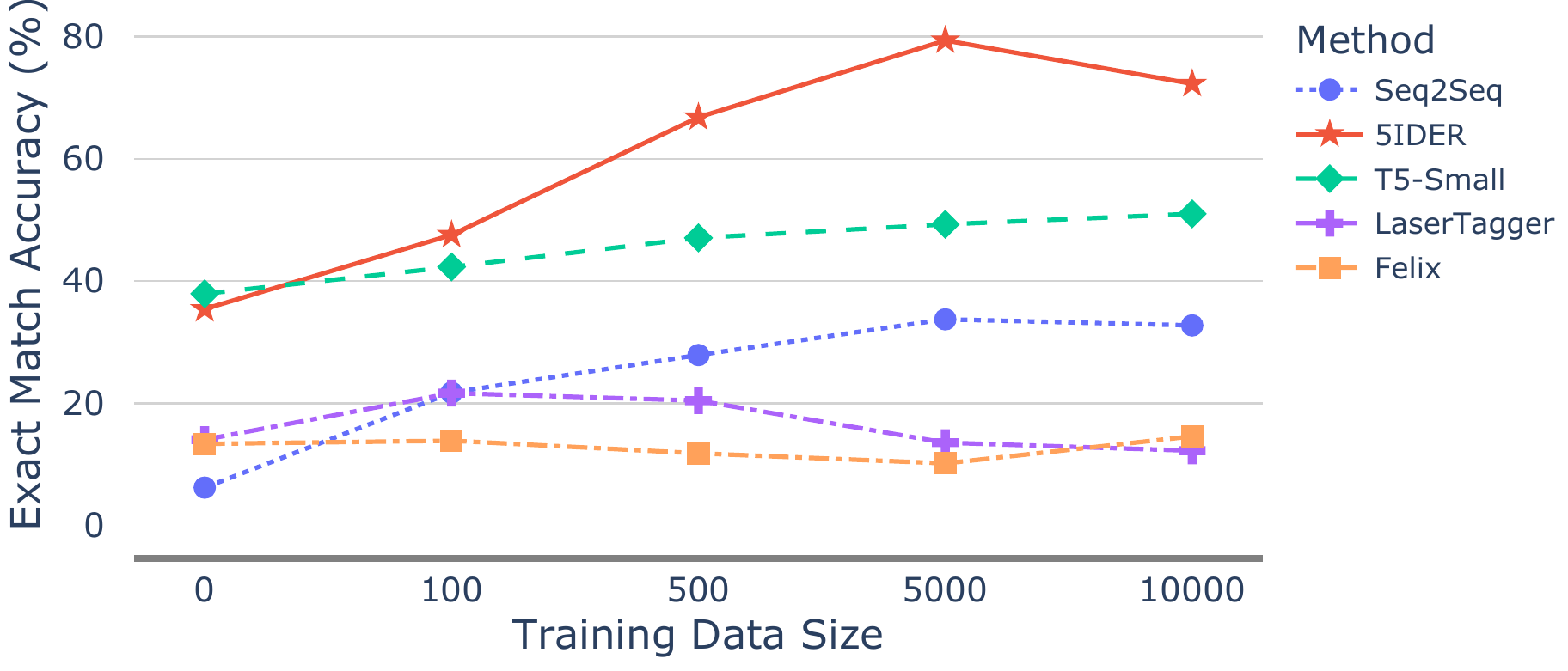}
  \caption{Performance of various approaches on use-case composition. The x-axis shows training data size (in number of data points), while the y-axis shows Exact Match Accuracy (\%).}
  \label{fig:usecase_composition}
  \vspace{-1.25em}
\end{figure}

We also experimented with challenging compositional use cases; our experimental results are shown in  Figure~\ref{fig:usecase_composition}. All models are trained with multitask single use case data, with a mixture of compositional data with varying dataset sizes. The x-axis shows the amount of compositional data used, while the y-axis shows the corresponding exact match accuracy on the test set. Note that the set of templates used to generate the train and test sets are disjoint, which ensures that the task cannot be solved through simple memorization.

Due to the aforementioned limitations, we find that LaserTagger and Felix perform  relatively poorly irrespective of compositional data size. Interestingly, we find that 5IDER consistently outperforms the Seq2Seq model: in particular, 5IDER with no compositional training data outperforms the best Seq2Seq model trained with 5k data points. This indicates that 5IDER generalizes substantially better than the baseline systems on compositional use cases, without additional training data.

In addition, 5IDER's zero-shot use case composition performance is very close to that of the much larger T5 model. With 100 training data points, 5IDER achieves performance competitive to that of the best T5 model, and with just 500 training data points, 5IDER significantly outperforms the T5 model trained with 10k data points, using 1/20\textsuperscript{th} the data, showcasing its great data efficiency on compositional use cases.

\section{Conclusion}

This work proposed a generalizable, multitasking, non-autoregressive query rewriting framework that handles 5 conversational use cases and their combinations. This model showed competitive performance in each use case, and significantly outperformed a fine-tuned T5-Small model in use case composition, while being 15 times smaller and 25 times faster.

\section{Acknowledgements}

The authors would like to thank Woojay Jeon, Leon Zhang, Dhivya Piraviperumal, Yuan Zhang, Nidhi Rajshree and Anuradha Gali for their help, support and feedback, and the anonymous reviewers for their helpful comments and suggestions.

\FloatBarrier

\bibliographystyle{IEEEtran}

\bibliography{interspeech23}

\end{document}